\documentclass[letterpaper, 10 pt, conference, onecolumn]{ieeeconf}  

\IEEEoverridecommandlockouts                              

\usepackage{float}
\usepackage{subfigure}

\usepackage{graphicx}
\usepackage{amssymb,amsmath}
\usepackage{bm} 
\usepackage{stackrel}
\usepackage{mathtools}
\usepackage{algorithm}
\usepackage{algorithmic}

\DeclarePairedDelimiter{\ceil}{\lceil}{\rceil}
\DeclarePairedDelimiter{\floor}{\lfloor}{\rfloor}
\DeclareMathOperator*{\argmax}{arg\,max}
\DeclareMathOperator*{\argmin}{arg\,min}
\usepackage{xcolor}
\newtheorem{remark}{Remark}
\newtheorem{definition}{Definition}
\newtheorem{lemma}{Lemma}
\newtheorem{theorem}{Theorem} 

\newtheorem{assumption}{Assumption}
\newtheorem{proofoflemma}{Proof of Lemma}
\newtheorem{proofoftheorem}{Proof of Theorem}
\title{\LARGE \bf A Decentralized Policy with Logarithmic Regret for a Class of Multi-Agent Multi-Armed Bandit Problems with Option Unavailability Constraints and Stochastic Communication Protocols}
        
\author{Pathmanathan\,Pankayaraj$^{1}$, D. H. S. Maithripala$^{2}$, and J. M. Berg$^{3}$
\thanks{$^{1}$Intern, Center for Research and Innovation Services, Sri Lanka Technological Campus, Padukka, CO 10500, Sri Lanka
        {\tt\small pankayarajp@sltc.ac.lk}}%
\thanks{$^{2}$ Department of Mechanical Engineering, University of Peradeniya, KY 20400, Sri Lanka / Sri Lanka Technological Campus, Padukka, CO 10500, Sri Lanka
        {\tt\small smaithri@eng.pdn.ac.lk}}%
\thanks{$^{3}$ US National Science Foundation, 2415 Eisenhower Ave, Alexandria, VA 22314, USA
        {\tt\small jberg@nsf.gov}}
}

\begin{document}

\maketitle
\thispagestyle{empty}
\pagestyle{empty}

\allowdisplaybreaks

\begin{abstract}
This paper considers a multi-armed bandit (MAB) problem in which multiple mobile agents receive rewards by sampling from a collection of spatially dispersed stochastic processes, called \textit{bandits}. The goal is to formulate a decentralized policy for each agent, in order to maximize the total cumulative reward over all agents, subject to option availability and inter-agent communication constraints. The problem formulation is motivated by applications in which a team of autonomous mobile robots cooperates to accomplish an exploration and exploitation task in an uncertain environment. Examples might include prospecting for and collecting mineral resources, or finding and marking unexploded ordnance. Bandit locations are represented by vertices of the \textit{spatial graph}. At any time, an agent's option consist of sampling the bandit at its current location, or traveling along an edge of the spatial graph to a new bandit location. Communication constraints are described by a directed, non-stationary, stochastic \textit{communication graph}. At any time, agents may  receive data only from their communication graph in-neighbors. 
The cumulative loss of reward due to sampling of sub-optimal options is called \textit{regret}. 
For the case of a single agent on a fully connected spatial graph, it is known that the expected regret for any optimal policy is necessarily bounded below by a function that grows as the logarithm of time. A class of policies called upper confidence bound (UCB) algorithms asymptotically achieve logarithmic regret for the classical MAB problem. In this paper, we propose a UCB-based decentralized motion and option selection policy and a non-stationary stochastic communication protocol that guarantee logarithmic regret. To our knowledge, this is the first such decentralized policy for non-fully connected spatial graphs with communication constraints. When the spatial graph is fully connected and the communication graph is stationary, our decentralized algorithm matches or exceeds the best reported prior results from the literature. The paper shows how the performance of the algorithm depends on parameters such as the total number of agents, the maximum number of in-neighbors, and the mobility of the agents.
\end{abstract}

\section{Introduction} \label{sect:Introduction}

This paper considers a multi-armed bandit (MAB) problem in which multiple mobile agents receive rewards by sampling from a collection of spatially distributed stochastic processes, with the goal of maximizing the total cumulative reward of all agents. Each agent is allowed to communicate only with a limited number of neighboring agents. The problem formulation is motivated by applications in which a team of autonomous mobile robots cooperates to accomplish a spatially distributed task in an uncertain environment. Examples might include prospecting for and collecting mineral resources, or finding and marking unexploded ordnance.  Movement constraints on the agents are described by a connected spatial graph with vertices representing the sampling options, and with edges representing directly connected options and the distances between those options. Communication connectivity between agents is described by a second non-stationary stochastic graph, which describes the subset of agents from which data may be received at a particular time. The limited communication makes it infeasible to direct all agents from a central location with perfect knowledge of the entire sampling history. Instead, each agent implements a decentralized policy, based only on local information.  

It is known that optimal policies for MAB problems must intersperse \textit{exploitation} -- that is, repeated sampling of the option with the highest expected reward -- with \textit{exploration} -- that is, sampling of options with lower expected rewards in order to reduce uncertainty about their distributions \cite{Gittins,Sutton,Robbins,LaiRobbins}. The loss of reward due to exploration is called \textit{regret}. In the standard MAB problem minimizing regret is equivalent to maximizing reward. For the case of a single agent on a fully connected spatial graph, it is known that the optimal expected regret is asymptotically bounded below by a function that grows as the logarithm of time \cite{LaiRobbins}. A class of policies called upper confidence bound (UCB) algorithms asymptotically achieve logarithmic regret for the classical MAB problem \cite{AgrawalSimpl,Auer,Kauffman,Garivier2011B,Kaufmann2016,Reverdy}. 

The class of problems we consider in this paper fall under the broad category of  decentralized multi agent multi armed bandit problems (D-MAMAB) \cite{Kalathil,LandgrenECC,Kolla,LandgrenCDC,Patrick}. A running consensus, with agents observing the estimates of other agents through communication, is employed in \cite{LandgrenECC,LandgrenCDC,Patrick}. Fixed graphs as well as stationary stochastic communication graphs are considered in these studies.
A brief review of the existing D-MAMAB schemes that use a running consensus of the estimates is provided in \cite{Patrick}. They also propose a novel running consensus based algorithm that results in a lower regret than the other existing consensus based algorithms. In contrast the work by \cite{Kolla} considers a D-MAMAB scheme where the agents use a fixed communication graph to communicate only the instantaneous rewards obtained by the agents. This ideas is extended to the case of an independently and identically distributed Erd\H{o}s-R\'enyi (ER) graph based stochastic communication policy in \cite{LandgrenCDC,UdariNaomi}.

In this paper, we propose a UCB-based movement and option selection policy and a non-stationary stochastic communication protocol which, to our knowledge, is the first decentralized policy guaranteeing asymptotically logarithmic regret for non-fully connected spatial graphs with communication constraints. When the spatial graph is fully connected and the communication graph is stationary, our decentralized algorithm matches or exceeds the best reported prior results from the literature. The communication protocol requires transmission only of the agent's most recent measurement. Therefore communication bandwidth requirements are moderate, and no storage of past measurements is required. The paper considers various ways in which performance may be improved, including the dependence on parameters such as the total number of agents, the number of agents communicating at a given time, and the mobility of the agents. 

The UCB based based policy that we propose utilizes a reward estimate that depends both on the option rewards obtained by the agent itself as well as those communicated by its neighbors. Redundancies in choice, due to two or more neighbors selecting the same option, are disregarded in the estimation. Each agent maintains an estimate of the rewards obtained by all other agents solely depending on the communicated rewards. Thus these estimates represent, in a certain sense, the belief that each agent has of the estimates that the other agents make.  
The paper proves that this policy guarantees a logarithmically bounded self regret irrespective of the communication strategy that the agents choose. In the proposed UCB based communication strategy the agents communicate with other agents who they believe to are most likely to be exploring. This in contrast to running consensus schemes, where the agents attempt to synchronize their estimates, the approach used here allows communication to be fully utilized for exploration. A preliminary version of these results on a fully connected spatial graph was presented in \cite{PankayarajECC}.

In section \ref{Secn:MAMABProblem} we present the notations and a precise definition of the class of D-MAMAB problems that is considered in this paper. Section \ref{Secn:RegretAnalysis} proposes the novel UCB based policy. This section also proves that the policy guarantees logarithmic regret. It also shows that the regret reduces with increasing connectivity. The effectiveness of the proposed scheme is demonstrated through simulations in section \ref{Secn:Simulations}.

\section{The Constrained Decentralised Multi Agent Multi Armed Bandit Problem}\label{Secn:MAMABProblem}
This section defines the Decentralised Multi Agent Multi Armed Bandit (D-MAMAB) problem that is considered in this paper. At any given time an agent will only have access to a limited set of options.  The options will be assumed to be distributed along the vertices of a fixed connected graph and the agents will be allowed to move from one vertex to another along the edges of the graph. At a given time step an agent can only move along one edge of the graph. An agent is only allowed to pick the option that is located at the vertex it occupies.
The agents may choose which vertex to move to, sample or not the option at the vertex it occupies, and choose which agents to communicate with. When agents communicate with each other they only share the value of the option it has sampled at that time instance. Such a choice that does not depend on any global information is called a \emph{decentralized policy}. The MAMAB problem that we will deal with in this paper is that of designing a suitable decentralized policy that maximizes the reward obtained by every agent. This will also ensure that the total network reward is maximized. In what follows we will provide a formal statement of this objective.

Let the total number of agents be denoted by $n_A$ and indexed by the set $\{1,2,\cdots,n_A\}$. The options and the reward associated with each option are assumed to satisfy the following explicit assumptions:
\begin{assumption}\label{as:MABassumption}\mbox{}
\begin{description}
\item[(a)] The options are taken to be distributed over the nodes of a fixed, connected, undirected graph $\mathcal{G}^o=(\mathcal{V}^o,\mathcal{E}^o)$. 
Given two options $i,j\in \mathcal{V}^o$ denote by $d(i,j)$ the shortest distance from $i$ to $j$ along the edges of the graph. For convenience we will assume that the distance between any two immediate neighbors of the undirected graph $\mathcal{G}^o$ is one unit (ie. the distance associated with any edge of the graph is one). It will also be assumed that the graph $\mathcal{G}^o=(\mathcal{V}^o,\mathcal{E}^o)$ is known to every agent. 
\item[(b)] A well defined optimal option  $i_*$ exists. That is, there exists an $i_*\in\mathcal{V}^o $ and $\Delta, \bar{\Delta}>0$ such that $\Delta\leq E({X}_{i_*}^\nu )-E(X_{i}^r)\leq \bar{\Delta}$ for all $r,\nu>0$ and $i\in\mathcal{V}^o$ that satisfy $i \neq i_{*}$.
\item[(c)] The reward associated with each option $i\in\mathcal{V}^o$ is given by a, possibly non stationary, stochastic process $\{X_i^t\}$ where $X_i^t$ is a sub Gaussian random variable and hence satisfies
\begin{align*}
\mathbb{E}\left(e^{\lambda X_i^t}\right)\leq e^{\lambda E(X_i^t)+\frac{\lambda^2{\sigma}^2}{8}},
\end{align*}
for some $\sigma>0$ and every $\lambda>0$. 
\end{description}
\end{assumption}
\vspace{0.25cm}
\noindent The last of the the above conditions imply that the results derived here are also valid for any stochastic process that takes values in a  bounded interval of length $\sigma$.


The random variable $\varphi_k^t$ corresponds to the option that is chosen by agent $k$ at time $t$. At time $t$ a given agent $k$ is at some vertex $n_k^{t}\in \mathcal{V}^o$ and it will decide to choose the option at $n_k^{t}$ or to refrain from choosing it. In the latter instance we set $\varphi_k^{t}=\varnothing$. At the time step $t$ it will also decide on a set of agents $\mathcal{N}_k^{t}\subseteq \{1,2,\cdots,n_A\}$ to communicate with.  \emph{The only information that an agent receives from another agent is the option that it has chosen, $\varphi_k^t$, and the corresponding reward, $X_{\varphi_k^t}^t$.} For mathematical completeness we set $X_{\varphi_k^t}^t\equiv 0$ when $\varphi_k^t=\varnothing$.
The resulting communication graph is allowed to be directed.  The discrete random variable $\mathcal{N}_{k}^t$  will be referred to as the \emph{neighbors of agent $k$ at time $t$}. By convention we let $k\in \mathcal{N}_{j}^t$. Let $\mathcal{N}_{k\alpha}$ be a subset of $\left\{1,2,\cdots,n_A\right\}$ that contains $k$ and let 
$\mathcal{N}_{kP}$ be the space of all such subsets of $\left\{1,2,\cdots,n_A\right\}$ (there are a total number of $|\mathcal{N}_{kP}|=2^{(n_A-1)}$ such sets). Then the discrete random variable $\mathcal{N}_k^t$ takes values in the set $\mathcal{N}_{kP}$. Based on the information it has obtained at time $t$, either by sampling the options on its own or through communication, the agent $k$ may decide to move to another vertex $n_k^{t+1}\in \mathcal{V}^o$ that is in the neighborhood of $n_k^{t}$ at time $t+1$. If so desired the agent is also allowed to remain at the vertex it occupies. A neighborhood of an option $i\in \mathcal{V}^o$ will be denote by $\mathcal{I}_{i}\subseteq \mathcal{V}^o$. 
The triple $\pi_j^t\triangleq (n_j^t,\varphi_j^t,\mathcal{N}_j^t)$ is the outcome of a suitable policy that will only depend on the information available to agent $j$ at time $t-1$.
Let
$\mathcal{F}^{t}_j$ be the sigma algebra generated by the random variables 
$\{(\mathcal{N}_j^\nu,n_j^\nu,\{X_{\varphi_k^\nu}^\nu\}_{k\in \mathcal{N}_j^\nu})\}_{\nu=1}^t$ and $\mathcal{F}^1_j\subset \mathcal{F}^2_j\subset \cdots \mathcal{F}^{t}_j$ be the corresponding filtration. Then the policy $\pi_j^t$ is a $\mathcal{F}^{t-1}_j$ measurable random variable.
\begin{definition}
The Decentralized Multi Agent Multi Armed Bandit (D-MAMAB) problem that we solve in this paper is that of finding a suitable decentralized policy $\pi_j^t=(n_j^t,\varphi_j^t,\mathcal{N}_j^t)$ for each agent so that the expectation of the cumulative reward obtained by each agent is maximized. The policy $\pi_j^t$ will be based solely on the information available to the agent at the previous time step $t-1$ and hence is a $\mathcal{F}^{t-1}_j$ measurable random variable. The solution will be subject to the conditions stated in assumption \ref{as:MABassumption}.
\end{definition}

\begin{figure}[H]
		\centering
		\begin{tabular}{c}
		\includegraphics[scale = 0.45]{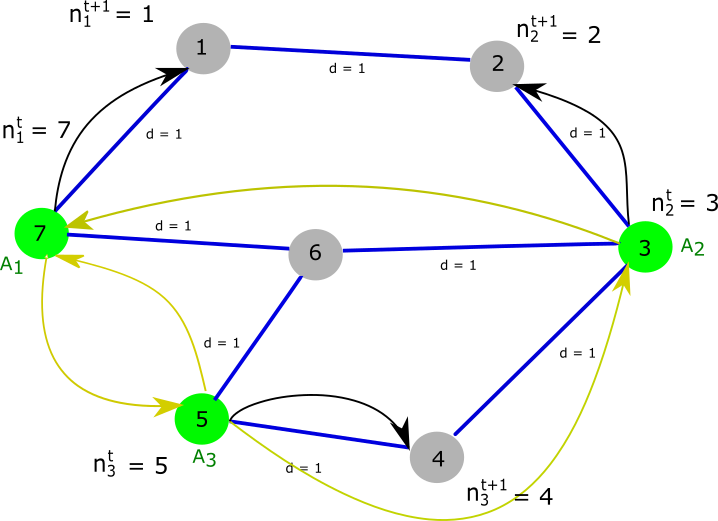}  
		\end{tabular}
		\caption{Example of the problem considered in the paper.}
		\label{Fig:Example}

\end{figure}

We pause for a while to illustrate the problem considered in this paper using a simple example. The example is for a seven option three agent MAMAB problem. It is represented using figure \ref{Fig:Example}. Let  $\mathcal{G}^o=(\mathcal{V}^o,\mathcal{E}^o)$ be the fixed un-directed graph shown in this figure. The vertices $\mathcal{V}^o=\{1,2,\cdots,7\}$ are shown by circles while the edges $\mathcal{E}^o$ are indicated by the blue lines in figure \ref{Fig:Example}. 
The seven options of the bandit are taken to be distributed along the vertices of the graph $\mathcal{G}^o=(\mathcal{V}^o,\mathcal{E}^o)$ while the edges  indicate the paths that an agent can follow. The distance between any two adjacent nodes of the graph is taken to be $d=1$.  The agents are labelled $\{A_1,A_2,A_3\}$ and the nodes they occupy at time $t$ are highlighted by the color green. Thus we have that $n_1^t=7, n_2^t=3,n_3^t=5$.
Based on the information an agent has, that is made available solely through exploration or communication with neighbors up to time $t$,  it will decide at time $t+1$ to move to a neighbor of the option that it occupied at $t$. 
For instance at the time step $t+1$ agent $A_2$ can only move to the options in the neighborhood of the option $3$ that is given by the set $\mathcal{I}_{3}=\{2,4,6\}$. The black arrows in this example indicate the node that each agent has decided to move. Accordingly we see that $A_2$ has decided to move to option 2 at time $t+1$ (ie. $n_2^{t+1}=2$). The agents then decide whether or not to sample the option that is located at the node it had moved to and which agents to communicate with based on the information it had at $t$. 
If agent $A_2$ decides to choose option 2 at time $t+1$ we set $\varphi_2^{t+1}=2$ or if otherwise we set $\varphi_2^{t+1}=\varnothing$.  
The yellow arrowed lines denote the directed edges of the communication graph at time $t$. Thus, for instance, according to our notation we have $\mathcal{N}_1^t=\{1,3\}$, $\mathcal{N}_2^t=\{2,1\}$ and $\mathcal{N}_3^t=\{3,1,2\}$.


The indicator random variable corresponding to an event $\mathcal{M}$ will be denoted by $ \mathbb{I}_{\mathcal{M}}$. Then 
$\mathbb{I}_{\{\varphi_k^t=i\}}$ represent if the option $i$ was picked by agent $k$ at time $t$ and hence is a $\mathcal{F}^{t-1}_k$ measurable Bernoulli random variable. The cumulative  reward obtained by agent $j$ by sampling option $i$ in the time horizon $[1,T]$ is defined by
$S^s_{ij}(T)\triangleq \sum_{t=1}^TX_i^t\mathbb{I}_{\{\varphi_{j}^t=i\}}$.
Then the total reward obtained by agent $j$ is $S^s_{j}(T)\triangleq \sum_{i\in \mathcal{V}^o}S^s_{ij}(T)$ and the total reward of the network is  
$S^s(T)\triangleq \sum_{j=1}^{n_A}\sum_{i\in \mathcal{V}^o}S^s_{ij}(T)$.

In the standard single agent MAB problem maximizing the reward is equivalent to minimizing the regret. Each time an agent adds the reward of a suboptimal option to its collected reward it also accumulates a regret that is equal to the difference between the optimal reward and the added suboptimal reward. Therefore the total cumulative self regret of the $j^\mathrm{th}$ agent, due to the sub optimal option $i$ being chosen by $j$ within the time horizon $T$ is defined to be 
\begin{align*}
R^s_{ij}(T)&\triangleq \mathbb{E}\left(\sum_{t=1}^T\left(X_{i_*}^t-X_i^t\right)\mathbb{I}_{\{\varphi_{j}^t=i\}}\right)\leq \bar{\Delta}\mathbb{E}\left(\sum_{t=1}^T\mathbb{I}_{\{\varphi_{j}^t=i\}}\right).
\end{align*}
Here the $\mathcal{F}^{t-1}$ measurable random variable 
\begin{align*}
N^s_{ij}(t)\triangleq \sum_{\nu=1}^t\mathbb{I}_{\{\varphi_{j}^\nu=i\}},
\end{align*}
denotes the total number of times that agent $j$ has chosen $i$ in the time horizon $[1,t]$.
Then the above expression for the regret takes the form $R^s_{ij}(T)\leq \bar{\Delta}\mathbb{E}\left(N^s_{ij}(T)\right)$. This says that the individual agent self regret per non optimal option is bounded above by the expectation of the number of times that the agent has sampled the  suboptimal  option in the time horizon $[1,T]$. 
The total regret accrued by agent $j$ is simply the summation of $R^s_{j}(T)\triangleq \sum_{i\neq i_{*}}R^s_{ij}(T)$ over all suboptimal options and the total network regret is obtained by summing this over all agents $R^s(T)\triangleq \sum_{j=1}^{n_A}\sum_{i\neq i_{*}}R^s_{ij}(T)$. Hence, in order to simplify the notation, it suffices to only consider the regret per suboptimal option per agent where the total network values can be obtained by summing over all suboptimal options and agents.

The $\mathcal{F}^{t-1}_j$ measurable Bernoulli random variable 
\begin{align}
\epsilon_{ij}^t\triangleq  \left\{
	\begin{array}{cl} 1 & \:\:\mathrm{if}\:\:\:\left(\sum_{k\in \mathcal{N}_j^t}\mathbb{I}_{\{\varphi_k^t=i\}}\right)\neq 0\\
  	0 & \:\:\: {\mathrm{o.w.}}\end{array}\right.,\label{eq:epsijt}
\end{align}
contains the information whether $j$ has received information of $i$ being picked either by itself or by one of its neighbors at time $t$.  
Typically the policy $\pi_j^t=(n_j^t,\varphi_j^t,\mathcal{N}_j^t)$ depends on the estimates that one makes of $\mathbb{E}\left(X_i^t\right)$.
The best possible approach to estimating the conditional expectation of $X_i^t$ is to use the full information one has access to. One such approach is to use the sample mean estimate, $\widehat{X}_i^t$, that is defined by
\begin{align}
N_{ik}(t)&\triangleq  \sum_{\nu=1}^t\epsilon_{ik}^\nu=N_{ik}({t-1})+\epsilon_{ik}^t,\label{eq:Nijt}\\
\widehat{X}_{ik}^t&\triangleq 
\frac{1}{N_{ik}(t)}\left(\sum_{\nu=1}^tX_i^\nu \epsilon_{ik}^\nu \right)=\frac{1}{N_{ik}(t)}\left(N_{ik}(t-1)\widehat{X}_{ik}^{t-1}+X_i^t\epsilon_{ik}^t\right),\label{eq:SampleMean}
\end{align}
We set ${X}_{ik}^0=\bar{x}_{ik}$ and $\widehat{\mu}_{ik}^0=\bar{x}_{ik}$ and $N_{ik}^0=1$ for all $i,k$ where $\bar{x}_{ik}$ are generated from some distribution  representing a prior belief of the options. The random variable $\widehat{X}_{ik}^t$ is the conditional sample mean of $X_i^t$ that is estimated by the agent $k$.
The random variable 
$N_{ij}(t)$
is the $\mathcal{F}_j^{t-1}$ measurable random variable that denotes the total number of times that $j$ has received information of $i$ being picked either by itself or by one of its neighbors at time $t$. The random variable $N^c_{ij}(t)\triangleq \sum_{\nu=1}^t\mathbb{I}_{\{\varphi_{j}^\nu\neq i\:\&\:\epsilon_{ij}^\nu=1\}}$ represents the number of times that agent $j$ has received information of option $i$ purely through means of communication.


A Hoeffding type tail bound is provided in Theorem 4 of \cite{Garivier2011B} for the random summand of pre-visible random variables. Setting $X_t=X_r^t$, $Y(t)=Y_{rk}(t)\triangleq \sum_{t=1}^T\left(X_r^t-\mathbb{E}\left(X_r^t\right)\right)\epsilon_{rk}(t)$, $\epsilon_t=\epsilon_{rk}(t)$, and $N(t)=\sum_{\tau=1}^t\epsilon_{rk}^\tau$ in this result it follows that the sample mean estimator defined  by (\ref{eq:Nijt}) -- (\ref{eq:SampleMean}) satisfies the tail bound 
{\small
\begin{align}
\mathcal{P}\left(\left\{\left|\widehat{X}_{rk}^t- \widehat{\mu}_{rk}^t\right|>\delta_r\sqrt{\frac{\Phi(t)}{N_{rk}(t)}}\right\}\right)\leq 2\ceil*{\frac{\ln{t}}{\ln{(1+\eta)}}} \,\exp{\left(-\frac{2\delta_r^2\Phi(t)}{{\sigma}^2(1+\eta)^{\frac{1}{2}}}\right)},\label{eq:GarivierHoeffding}
\end{align}
}
where $\eta>0$ is a constant, $\Phi(t)$ is some positive function of $t$ and  $\widehat{\mu}_{rk}^t$ is defined by 
\begin{align*}
\widehat{\mu}_{rk}^t&\triangleq 
\frac{1}{N_{rk}(t)}\left(\sum_{\nu=1}^t\mathbb{E}\left(X_r^\nu \right)\epsilon_{rk}^\nu \right).
\end{align*}

Expression (\ref{eq:GarivierHoeffding}) represents the belief that the agent $k$ has of the estimate of the reward of option $r$. The number of times that agent $k$ has become aware of $r$ being chosen depends on the policy $\pi_k^t=(n_k^t,\varphi_k^t,\mathcal{N}_k^t)$. 
The right hand side says that if the confidence on the belief of the estimate is to increase as $(1-2\vartheta/t^2)$, where $\vartheta =1/\sqrt{1+\eta}$, then the function $\Phi(t)$ 
must satisfy $ ({\sigma}^2\sqrt{1+\eta}/\delta_r^2)\log{\left(t\sqrt{\log{(t)}}\right)}\leq \Phi(t)$.
On the other hand since
\begin{align*}
\left\{\left|\widehat{X}_{rk}^t- \widehat{\mu}_{rk}^t\right|>\delta_r \right\}\subseteq \left\{\left|\widehat{X}_{rk}^t- \widehat{\mu}_{rk}^t\right|>\delta_r\sqrt{\frac{\Phi(t)}{N_{rk}(t)}}\right\}
\end{align*}
if and only if $N_{rk}(t)\geq \Phi(t)$, by choosing $\Phi(t)=({\sigma}^2\sqrt{1+\eta}/\delta_r^2)\log{\left(t\sqrt{\log{(t)}}\right)}$ the following lemma follows.
\begin{lemma}\label{lemma:GarivierHoeffding}
If $N_{rk}(t)\geq ({\sigma}^2\sqrt{1+\eta}/\delta_r^2)\log{\left(t\sqrt{\log{(t)}}\right)}$ then
\begin{align}
\mathcal{P}\left(\left\{\left|\widehat{X}_{rk}^t- \widehat{\mu}_{rk}^t\right|>\delta_r\right\}\right)\leq  \frac{2\vartheta}{t^2}.\label{eq:GarivierHoeffding2}
\end{align}
\end{lemma} 
This in essence implies that if an agent receives information from an option at least logarithmically often then it will be $(1-2\vartheta/t^2)$ confident that the estimate satisfies $\left|\widehat{X}_{rk}^t- \widehat{\mu}_{rk}^t\right|\leq \delta_r$.
Thus since $N_{rk}(t)=N_{rk}^s(t)+N_{rk}^c(t)$ an agent can increase the confidence that it has of $\widehat{X}_{rk}^t$ by means of effective communication.

\section{UCB Based Decentralized Policy}\label{Secn:RegretAnalysis}
In this section we propose a decentralized policy for the motion of the agents, $n_j^t$, the selection of options, $\varphi_j^t$, and the choice of neighbors to communicate with, $\mathcal{N}_j^t$, solely based on the information the agent has obtained up to the previous time step. That is, we will develop a $\mathcal{F}_j^{t-1}$ measurable policy $\pi_j^t=(n_j^t,\varphi_j^t,\mathcal{N}_j^t)$ for solving the D-MAMAB problem. The policy will be based on a set of UCB based cost functions that each agent maintains. Namely at each time instance $t$ the agents will (a) chose $n_j^t$, $\varphi_j^t$, and $\mathcal{N}_j^t$ based on the cost function values at $t-1$, and (b) update the cost function values.
We will first consider step of choosing $n_j^t$ and $\varphi_j^t$ that we will refer to as the \emph{UCB based motion allocation and option selection policy} and $\mathcal{N}_j^t$ the \emph{UCB based communication policy}.

\subsection{UCB Based Motion Allocation and Option Selection Policy}
Consider the stochastic process $\{\epsilon_{ij}(t)\}$ where $\epsilon_{ij}(t)$ is the $\mathcal{F}_j^{t-1}$ measurable Bernoulli random variable defined by (\ref{eq:epsijt}). 
For the MAB problem \cite{Auer} proposed a class of optimal policies popularly known as Upper Confidence Bound (UCB) policies that guarantee logarithmic regret for the standard MAB problem. These policies balance exploration and exploitation using a carefully chosen cost function. We use a slightly modified version of this cost function by introducing a distance penalty,
\begin{align}
Q_{ij}^t&\triangleq \widehat{X}_{ij}^t+\sqrt{\frac{(1+\alpha \bar{\tau})}{\left(1+\alpha\, d\left(n_j^t,i\right)\right)}\frac{\Psi_j(t)}{N_{ij}(t)}}.\label{eq:UCBQ}
\end{align}
Here $d\left(a,b\right)$ is the minimum distance between the nodes $a,b\in \mathcal{V}^o$ measured along the edges of the graph $\mathcal{E}^o$,  $\bar{\tau}\triangleq \max_{a,b\in \mathcal{V}^o}d(a,b)$ is the maximum distance between any two nodes of the graph $\mathcal{G}^o$ and $\alpha>0$ is a tuning parameter that can be used to enforce local exploration.
The term $\Psi_j(t)$ in the above cost function is logarithmically bounded from above and below and dictates the exploration or the uncertainty of the estimates made by the agents. The factor $1\leq \frac{(1+\alpha \bar{\tau})}{\left(1+\alpha\, d\left(n_j^t,i\right)\right)}\leq (1+\alpha \bar{\tau})$ is introduced to ensure that the cost function reduces with the distance to the option and hence encourages the agents to explore options that are nearby.
When $\Psi_j(t) \sim \log{(t)}$ and when the spatial graph is fully connected the above cost function reduces to the standard UCB based cost function proposed by \cite{Auer}.


\begin{definition}{\sl UCB Based Motion and Option Selection Policy}\\
Let 
\begin{align*}
\hat{n}_j^t&=\argmax_{i\in \mathcal{V}^o}\limits\,\{Q_{ij}^t\},
\end{align*}
where one chooses uniformly randomly when multiple choices exist. The \emph{UCB based motion and option selection policy $(\varphi_j^t,n_{j}^{t})$} is defined by
\begin{align}
n_j^{t+1}=\argmin_{n\in \mathcal{I}_{n_j^{t}}}\:\:d(n_j^{t},n)+d(n,\hat{n}_j^t).\label{eq:MotionAllocation}
\end{align}
	\begin{align}
	\varphi_j^{t+1}=\left\{
	\begin{array}{cl} \hat{n}_j^t & \:\:\mathrm{if} \:\:\:{n_j^{t+1}}=\hat{n}_j^t\\ %
  	\varnothing & \:\:\: {\mathrm{o.w.}}\end{array}\right.\label{eq:UCBAllocation}
	\end{align}	
\end{definition}

In the standard MAB problem the sole agent has access to all options at all times and hence $\mathcal{I}_{n_j^t}=\mathcal{V}^o$. This corresponds to a fully connected spatial graph. Then the policy reduces to just $\pi_j^t=\varphi_j^t$ and every agent picks some option at all time steps. This is in general not true in the case of the policy (\ref{eq:MotionAllocation}) -- (\ref{eq:UCBAllocation}) in the face of option unavailabilities.

\subsection{UCB Based Communication Policy}\label{Secn:UCBComPolicy}

An agent choses its neighbors to communicate with based on a UCB type selection rule. The rule depends only on the information the agent has. The policy we propose encourages agents to communicate with other agents who they believe are most likely to be exploring. We proceed to make this precise.

Let $\mathbb{I}_{\{j,k\}}^t$ denote the $\mathcal{F}_{t-1}$ measurable Bernoulli random variable that denotes if or not agent $j$ communicates with agent $k$. We do not require that the communication be bi-directional. That is we do not require $\mathbb{I}_{\{j,k\}}^t=\mathbb{I}_{\{k,j\}}^t$. Each agent maintains an estimate of the rewards estimated by other agents using only the information made available through communication. That is, we let $\widehat{X}_{ijk}^t$ be the estimate that $j$ makes of the estimate that $k$ has made of the reward of the option $i$ within the time horizon $[1,2,\cdots,t]$. Explicitly stated
\begin{align*}
\widehat{X}_{ijk}^t&\triangleq 
\frac{1}{N_{ijk}(T)}\left(\sum_{\nu=1}^TX_i^\nu\mathbb{I}_{\{j,k\}}^\nu\mathbb{I}_{\{\varphi_k^\nu=i\}}\right).
\end{align*}
Here
\begin{align*}
	N_{ijk}(t)&\triangleq N_{ijk}(t-1)+\mathbb{I}_{\{j,k\}}^t\mathbb{I}_{\{\varphi_k^t=i\}},
\end{align*}
is the random variable that denotes the number of times that $j$ has been made aware of by agent $k$ that it has chosen option $i$. Note that $N_{ijj}(t)=N_{ij}^s(t)$ and thus $\widehat{X}_{ijj}^t\neq \widehat{X}_{ij}^t$.
The definition below makes precise the UCB based novel communication policy that we propose in this paper.
\begin{definition}\label{def:UCB basedcommunication}
Let $\gamma_j$ be the maximum number of other agents that agent $j$ is allowed to communicate with and let
\begin{align}
	Q_{ijk}^t&\triangleq \widehat{X}_{ijk}^t+\sqrt{\frac{\Psi_{j}(t)}{N_{ijk}(t)}}\label{eq:UCBQc}
	\end{align} 
	where  $\Psi_{j}(t)$ satisfies $({\sigma}^2\sqrt{1+\eta})\log{\left(t\sqrt{\log{(t)}}\right)}\leq \Psi_{j}(t)$ for all $t>0$ and some $\eta>0$.
Define $\mathcal{Q}_{j}^t$ to be the  largest $\gamma_j$ values of the set $\cup_{k\neq j}\max \{Q_{ijk}^t\,|\,
\:\:i=1,2,\cdots,n_O\:\: s.t.\:\: i\neq \hat{n}_j^t\}$. 
When ambiguity arises due to repeated elements the ambiguity will be resolved by choosing in a uniformly random manner. 
Then agent $j$ chooses its neighbors according to the policy
\begin{align}
\mathcal{N}_j^{t+1}=\left\{\arg_{k} Q_{ijk}^t \:\:| \:\: Q_{ijk}^t\in\mathcal{Q}_{j}^t\right\}\cup\{j\}.\label{eq:UCB_Agent_allocation}
\end{align}
	\end{definition}
Note that (\ref{eq:UCB_Agent_allocation}) implies
	{\small
	\begin{align*}
	\mathbb{I}_{\{j,k\}}^{t+1}=\left\{
	\begin{array}{cl} 1 & \:\:\mathrm{if}\:\:\:Q_{ijk}^t\in\mathcal{Q}_{j}^t\\
  	0 & \:\:\: {\mathrm{o.w.}}\end{array}\right.
	\end{align*}}
\begin{remark}
This UCB based communication policy ensures that an agent communicates with other agents who the agent  believes to be most likely to be exploring than exploiting at that time instant.
\end{remark}

In the section below we will show that the agent regret remains logarithmically bounded if one uses the policy $\pi^t_j=(n_j^t,\varphi_j^t,\mathcal{N}_j^t)$ defined by (\ref{eq:MotionAllocation}), (\ref{eq:UCBAllocation}), and (\ref{eq:UCB_Agent_allocation}). In the simulations below we show that this policy significantly outperforms an i.i.d. ER graph random communication policy.
Notice that since the communication involves only two pieces of locally gathered information the scheme is easily scalable and is very communication-ally and computationally cost effective.

The indicator random variable $\mathbb{I}_{\{\varphi_{j}^t\neq i\:\&\:\epsilon_{ij}^t=1\}}$ tells us if $j$ has received information of option $i$ being picked by one of its neighbors when it has not sampled $i$. The number of times that this is true, $N^c_{ij}(t)\triangleq \sum_{\nu=1}^t\mathbb{I}_{\{\varphi_{j}^\nu\neq i\:\&\:\epsilon_{ij}^\nu=1\}}$, provides a measure of the effectiveness of communication. 
This value is always zero in the absence of communication and increases with increased communication. Thus we define the \emph{effect of communication} on agent $j$  to be
\begin{align}
C_{j}(T)&\triangleq \sum_{i\in \mathcal{V}^o} \frac{\mathbb{E}\left(N^c_{ij}(T)\right)}{\mathbb{E}\left(N^s_{ij}(T)\right)}
.\label{eq:CommunicationEfficiency}
\end{align}
This will be zero either in the absence of communication or when all the agents are behaving identically. A larger value for this index indicates improved communication effect.

\subsection{Regret Analysis}

The work of \cite{Auer} and the extension by \cite{Garivier2011B} to include non-stationary bandits show that the UCB based allocation rule, of choosing the option that corresponds to the maximal cost, guarantees that a logarithmic bound is achieved for the standard MAB problem. Crucial in the proof of this result is the observation that the probability of picking a suboptimal option when it has been picked more than the factor $\Psi_k(t)$ is bounded by the tail probabilities of the estimator. This is stated formally in the lemma below and is proven for the sake of completeness in the appendix by closely following the proof of \cite{LaiRobbins}.
\begin{lemma}\label{Lemm:Main}
Let the conditions of assumption \ref{as:MABassumption} hold. 
Then the UCB based motion selection and option allocation rule $(n_k^t,\varphi_k^t)$ given by (\ref{eq:MotionAllocation}) and (\ref{eq:UCBAllocation}) with $({\sigma}^2\sqrt{1+\eta}))\log{\left(t\sqrt{\log{(t)}}\right)}\leq   \Psi_{k}(t) \leq  \Psi{(t)}$ will ensure that for all $i\neq i_*$
{
\begin{align*}
\mathcal{P}\left({\left\{Q_{i_*k}^t< {Q_{ik}^t}\:\:\&\:\: N_{ik}(t)> \ceil*{\frac{4(1+\alpha \bar{\tau})}{{\Delta^2}}\,\Psi(t)}\:\: \&\:\:{i \neq i_*}\right\}}\right)
\leq \frac{4\vartheta}{t^2},
\end{align*}
}
where 
$\widehat{X}_{rk}^t$, $\widehat{\mu}_{rk}^t$, and $Q_{rk}^t$ are defined by (\ref{eq:Nijt}),(\ref{eq:SampleMean}) and (\ref{eq:UCBQ}) respectively. Furthermore this also implies that
\begin{align*}
\mathcal{P}\left(\left\{Q_{i_*k}^{t}<{Q_{ik}^{t}}\:\: \&\:\:{i \neq i_*}\right\}\right)
&\leq \frac{4\vartheta}{t^2}+\mathcal{P}\left(\left\{N_{ik}(t)\leq\,\ceil*{\frac{4(1+\alpha \bar{\tau})}{{\Delta^2}}\,\Psi(t)}\right\}\right).
\end{align*}
\end{lemma}

By closely following \cite{LaiRobbins} we also prove in the appendix that, irrespective of the communication strategy used, the UCB based motion and option allocation rule (\ref{eq:MotionAllocation}) -- (\ref{eq:UCBAllocation}) guarantees that the self regret is logarithmically bounded. 
\begin{theorem}\label{Theom:ijRegretDependent}
Let the communication graph process $\{\mathcal{G}^t\}$ be arbitrary. If the conditions of assumption \ref{as:MABassumption} hold and $\alpha,\eta>0$ then the UCB based motion selection and option allocation rule $(n_k^t,\varphi_k^t)$ given by (\ref{eq:MotionAllocation}) -- (\ref{eq:UCBAllocation}) with $({\sigma}^2\sqrt{1+\eta}))\log{\left(t\sqrt{\log{(t)}}\right)}\leq   \Psi_{k}(t) \leq  \Psi{(t)}$ for all $k\in\{1,2,\cdots,n_A\}$ will ensure that the self regret satisfies
{\small
\begin{align*}
R^s_{ik}(T)&\leq \bar{\Delta} \mathbb{E}\left(N^s_{ik}(T)\right)\leq  \bar{\Delta}\left(2\left(1-f_{ik}(T)\right)+4\vartheta+f_{ik}(T)\,\ceil*{\frac{4\left(1+\alpha\,\bar{\tau}\right)}{{\Delta^2}}\,\Psi(T)}\right)\leq  \bar{\Delta}\left(4\vartheta+\ceil*{\frac{4\left(1+\alpha\,\bar{\tau}\right)}{{\Delta^2}}\,\Psi(T)}\right)
\end{align*}
}
for all $i\neq i_*$ and $j$  where $\vartheta=1/\log (1+\eta)$,  and
\begin{align*}
f_{ik}(T)\triangleq \frac{\mathbb{E}\left(\sum_{t=2}^T\mathbb{I}_{\left\{N_{ik}(t-1)\leq\,l(t-1)\right\}}\right)}{\mathbb{E}\left(\sum_{t=2}^T\mathbb{I}_{\left\{N^s_{ik}(t-1)\leq\,l(t-1)\right\}}\right)}\leq 1.
\end{align*}
When there is no communication the factor $f_{ik}(T)=1$ and it reduces as the communication increases. 
\end{theorem}
From the proof of this theorem it also follows that 
$\mathbb{E}\left(N^s_{ik}(T)\right)\leq  \left(4\vartheta+\ceil*{\frac{4\left(1+\alpha\,\bar{\tau}\right)}{{\Delta^2}}\,\Psi(T)}\right)$. The expected number of times an agent becomes aware of a
given option being chosen, $\mathbb{E}\left(N_{ik}(T)\right)$, is clearly greater than or equal to $\mathbb{E}\left(N^s_{ik}(T)\right)$. In the appendix we show that following lemma holds as well.
\begin{lemma}\label{Theom:Nijt}
Let the communication graph process $\{\mathcal{G}^t\}$ be arbitrary. If the conditions of assumption \ref{as:MABassumption} hold  then the UCB based motion selection and option allocation rule $(n_k^t,\varphi_k^t)$ given by (\ref{eq:MotionAllocation}) -- (\ref{eq:UCBAllocation}) with $({\sigma}^2\sqrt{1+\eta}))\log{\left(t\sqrt{\log{(t)}}\right)}\leq   \Psi_{k}(t) \leq  \Psi{(t)}$ for all $k\in\{1,2,\cdots,n_A\}$ and for some $\eta>0$, will ensure that 
{\small
\begin{align*}
\mathbb{E}\left(N_{ij}(T)\right)&\leq  \left(\max_{k,t\leq T}\left\langle |\mathcal{N}_{j}^t|\right\rangle_{\mathcal{P}\left(\mathcal{N}_j^t\,|\,\varphi_{k}^t=i \right)}\right)\left(4\vartheta+\,\ceil*{\frac{4\left(1+\alpha\,\bar{\tau}\right)}{{\Delta^2}}\,\Psi(T)}\right)
\end{align*}
}
where
\begin{align*}
\left\langle | \mathcal{N}_j^t |\right\rangle_{\mathcal{P}\left(\mathcal{N}_j^t\,|\,\varphi_{k}^t =i\right)} 
&\triangleq\sum_{\mathcal{N}_{j\alpha}\in \mathcal{N}_{jP}}\mathcal{P}\left(\{\mathcal{N}_j^t=\mathcal{N}_{j\alpha}\}\,|\,\{\varphi_{k}^t=i\} \right)|\mathcal{N}_{j\alpha}|.
\end{align*}
\end{lemma}

 \begin{remark}
When the number of connections are restricted to $\gamma_j$ we have that
$\left\langle | \mathcal{N}_j^t |\right\rangle_{\mathcal{P}\left(\mathcal{N}_j^t\,|\,\varphi_{k}^t =i\right)} \leq \gamma_j+1$. On the other hand when the graph is an independent process, $\mathcal{P}\left(\mathcal{N}_j^t\,|\,\varphi_{k}^t =i\right)=\mathcal{P}\left(\mathcal{N}_j^t \right)$ and hence
$\left\langle |\mathcal{N}_j^t|\right\rangle\triangleq  \sum_{\mathcal{N}_{j\alpha}\in \mathcal{N}_{jP}}|\mathcal{N}_{j\alpha}|\mathcal{P}\left( \mathcal{N}_j^t=\mathcal{N}_{j\alpha}\right)$. 
\end{remark}


According to the policy (\ref{eq:MotionAllocation}) -- (\ref{eq:UCBAllocation}) it is very likely that an agent will not choose any option at every time instant. The estimates $\widehat{X}_{ik}^t$ nor the number of times an option is chosen, $N_{ik}(t)$, will be updated during a time interval where an agent does not receive any information of the options.  
Thus during such an interval the estimated optimal option $\hat{n}_k^t$ will not get updated as well. However the policy (\ref{eq:MotionAllocation}) -- (\ref{eq:UCBAllocation}) guarantees that, if $Q_{i_*k}^t<Q_{ik}^t$ then agent $k$ will receive information of option $i$ in at most $d(n_k^t,i)\leq \bar{\tau}$ time steps. That is $N_{ik}\left(t+d(n_k^t,i)\right)\geq N_{ik}(t)+1$. On the other hand since the numerator of the second term in (\ref{eq:UCBQ}) is logarithmically increasing it also follows that if $N_{ik}(t)$ remains constant then $Q_{i_*k}^\nu<Q_{ik}^\nu$ for some $\nu>t$. Thus we see that the policy (\ref{eq:MotionAllocation}) -- (\ref{eq:UCBAllocation}) guarantees that every $N_{ik}(t)$ is an increasing function of time. Furthermore since, at least every $ \bar{\tau}$ time steps, every agent must necessarily receive information of some option being picked it also follows that 
\begin{align}
 \Gamma_j(t)&\triangleq \sum_{i\in \mathcal{V}^o}N_{ij}(t)\geq \floor*{\frac{t}{\bar{\tau}}},\label{eq:Nistarjt}
\end{align}
and hence that
\begin{align*}
N_{i_*j}(T)= \Gamma_j(T)- \sum_{i\neq i_*}N_{ij}(T)\geq \floor*{\frac{T}{\bar{\tau}}}-
(n_O-1)\left(4\vartheta+\,\ceil*{\frac{4\left(1+\alpha\,\bar{\tau}\right)}{{\Delta^2}}\,\Psi(T)}\right)\left(\max_{i,k,t\leq T}\left\langle |\mathcal{N}_{j}^t|\right\rangle_{\mathcal{P}\left(\mathcal{N}_j^t\,|\,\varphi_{k}^t=i \right)}\right).
\end{align*}
This implies that there exists a $T_c$ such that $N_{i_*k}(t)\geq (4{\sigma}^2\sqrt{1+\eta}/\Delta^2)\log{\left(t\sqrt{\log{(t)}}\right)}$ and hence from lemma \ref{lemma:GarivierHoeffding} that
\begin{align*}
\mathcal{P}\left(\left\{\left|\widehat{X}_{i_*k}^t- \widehat{\mu}_{i_*k}^t\right|>\frac{\Delta}{2}\right\}\right)\leq  \frac{2\vartheta}{t^2},
\end{align*}
for all $t\geq T_c$. Following the proof of Theorem 21 of \cite{Kaufmann2016} it can also be show that asymptotically, $\mathbb{E}(N_{ik}(t))$ is logarithmically bounded from below. 
Following the proof of lemma \ref{Theom:Nijt} we also see that
 \begin{align*}
\mathbb{E}(N^c_{ij}(T))
&\leq \left(\max_k\mathbb{E}\left(N^s_{ik}(T)\right)\right)\left(\max_{k,t\leq T}\left\langle |\mathcal{N}_{j}^t-1|\right\rangle_{\mathcal{P}\left(\mathcal{N}_j^t\,|\,\varphi_{k}^t=i \right)}\right).
\end{align*}
Using this it can be easily shown that the communication effect is upper bounded by a factor of the expected connectivity of the communication graph as stated in the lemma  below.


\begin{lemma}\label{Theom:ComEffect}
Let the communication graph process $\{\mathcal{G}^t\}$ be arbitrary. If the conditions of assumption \ref{as:MABassumption} hold  then the UCB based allocation rule $(\varphi_k^t,n_k^t)$ given by (\ref{eq:MotionAllocation}) -- (\ref{eq:UCBAllocation}) with $({\sigma}^2\sqrt{1+\eta}))\log{\left(t\sqrt{\log{(t)}}\right)}\leq   \Psi_{k}(t) \leq  \Psi{(t)}$ for all $k\in\{1,2,\cdots,n_A\}$ and for some $\eta>0$, will ensure that the effect of communication satisfies
\begin{align*}
C_{j}(T)&\leq  \sum_{i\in \mathcal{V}^o}h_i(T)\left(\max_{k,t\leq T}\left\langle |\mathcal{N}_{j}^t|-1\right\rangle_{\mathcal{P}\left(\mathcal{N}_j^t\,|\,\varphi_{k}^t=i \right)}\right)
\end{align*}
where
\begin{align*}
h_i(T)\triangleq \frac{\max_k\mathbb{E}\left(N^s_{ik}(T)\right)}{\min_k\mathbb{E}\left(N^s_{ik}(T)\right)}.
\end{align*}
\end{lemma}
\begin{remark}
In the special case where the the graph, $\{\mathcal{G}^t\}$, is an i.i.d process 
\begin{align*}
C_{j}(T)&\leq  \left(\left\langle |\mathcal{N}_{j}^t|\right\rangle-1\right)\sum_{i\in \mathcal{V}^o}h_i(T).
\end{align*}
On the other hand if the connectivity of each node $j$ is restricted to $\gamma_j$ then $\max_{k,t}\left\langle |\mathcal{N}_{j}^t|-1\right\rangle_{\mathcal{P}\left(\mathcal{N}_j^t\,|\,\varphi_{k}^t \right)}\leq \gamma_j$ and hence
\begin{align*}
C_{j}(T)&\leq  \gamma_j\sum_{i\in \mathcal{V}^o}h_i(T).
\end{align*}
\end{remark}


\section{Simulations}\label{Secn:Simulations}\label{Secn:Simulations}
\begin{figure}[H]
		\centering
		\begin{tabular}{cc}
		\includegraphics[scale = 0.5]{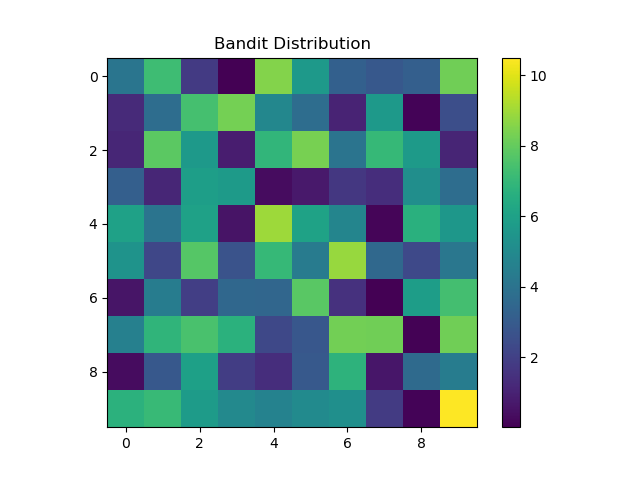} & \includegraphics[scale = 0.5]{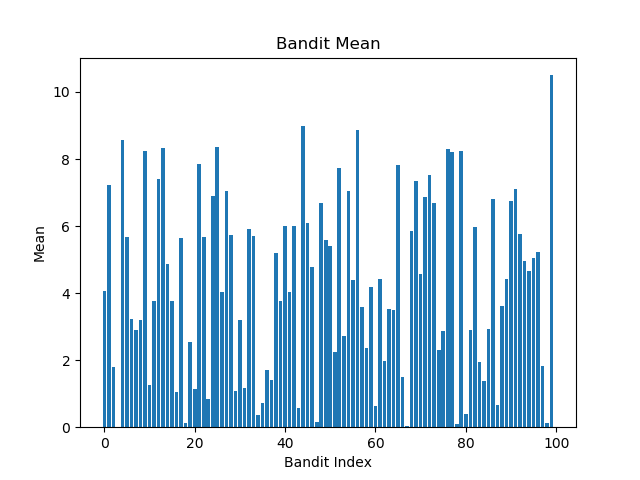}\\
		(a)   & (b) 
		\end{tabular}
		\caption{Options occupying each cell of a two dimensional lattice.}
		\label{fig:BanditMean}
\end{figure}

In this section we consider a set of 100 options located at the nodes of a $10\times 10$ two dimensional spatial lattice. The reward associated with each option is assumed to satisfy a Gaussian normal process with variance equal to 2. The intensity of the cell color depicts the size of the mean of the option where the brightest yellow corresponds to the option with the largest mean and the darkest blue corresponds to the option with the smallest mean. The option with the largest mean is the one that occupies the bottom right most corner cell that is highlighted in bright yellow in figure \ref{fig:BanditMean} (a). The figure \ref{fig:BanditMean} (b) shows the distribution of the numerical values of the means of each of the 100 options.

At the initial time step each agent $k$ initializes its  estimates, $\widehat{X}_{ik}^0$ by randomly sampling from a probability distribution that represents its prior belief of the option rewards. A time horizon of $T=20,000$ was chosen for each agent and the expectations were estimated by averaging over 20 trials. The total number of agents considered was $n_A=20$.

In the case where the communication graph $\mathcal{G}^t$ is an iid ER graph process, at each time $t$ the communication graph is an element of, $\mathbb{G}(n_A,p)$, the space of all possible ER graphs of $n_A$ nodes and edge probability $p$. In this case the probability of $j$ having the set of neighbors $\mathcal{N}_j^t=\mathcal{N}_{j\alpha}$ is given by
$ \mathcal{P}\left( \mathcal{N}_j^t=\mathcal{N}_{j\alpha}\right)=p^{|\mathcal{N}_{j\alpha}|-1}(1-p)^{n_A-|\mathcal{N}_{j\alpha}|}$.
Note that the expectation of the connectivity of the ER graph is given by $\langle |\mathcal{N}_{j\alpha}|\rangle-1 =(n_A-1)p$. In the case of the UCB based communication policy we restrict the connectivities to a certain fixed value $\gamma$. 

\begin{figure}[H]
	\centering	
	\begin{tabular}{cc}
	\includegraphics[scale = 0.315]{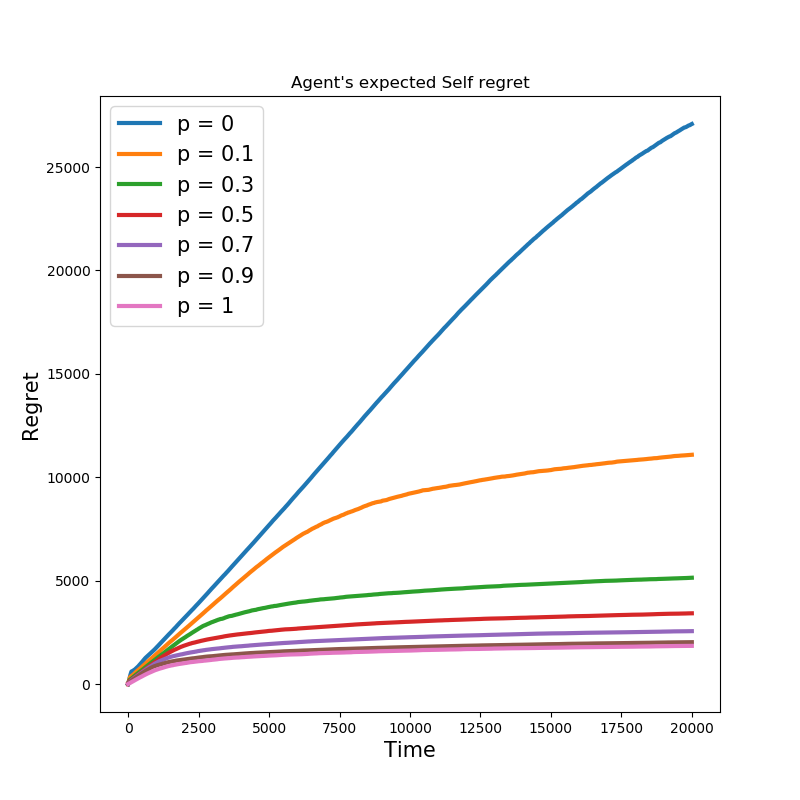} &
	\includegraphics[scale = 0.315]{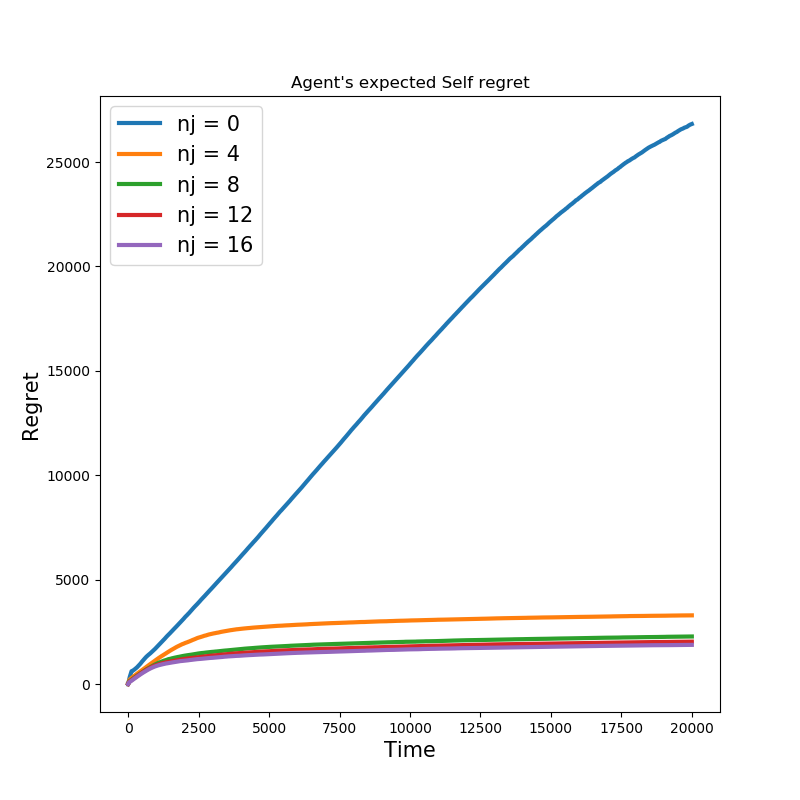}\\
	(a) ER Graph based communication & (b) UCB based communication
	\end{tabular}		
	\caption{The expected values of the self regret, $R^s_{ij}(T)$, 
	for several connectivity values of the communication graph for the: (a) iid ER graph communication policy, (b) UCB based communication policy (\ref{eq:UCB_Agent_allocation}).}	\label{fig:ERRegret}
	\end{figure}

\begin{figure}[H]
	\centering	
	\begin{tabular}{cc}
	\includegraphics[scale = 0.275]{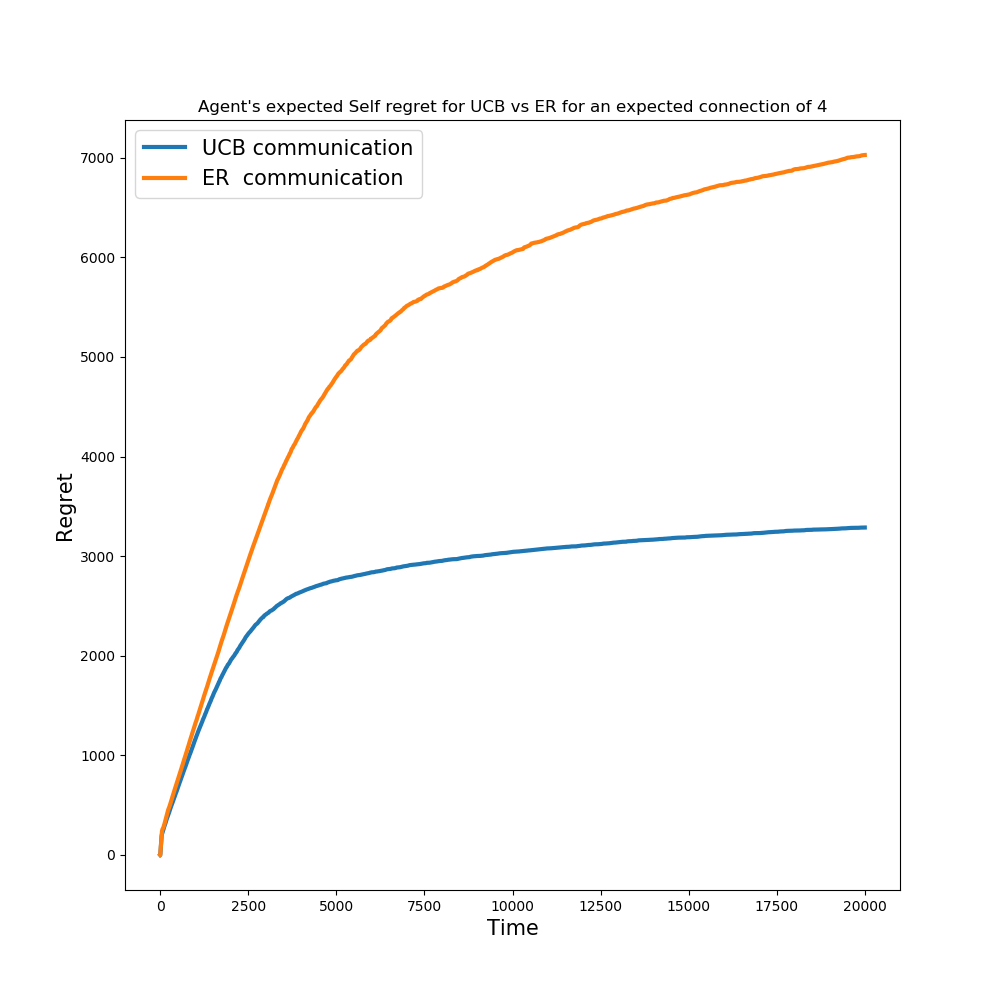} &
	\includegraphics[scale = 0.275]{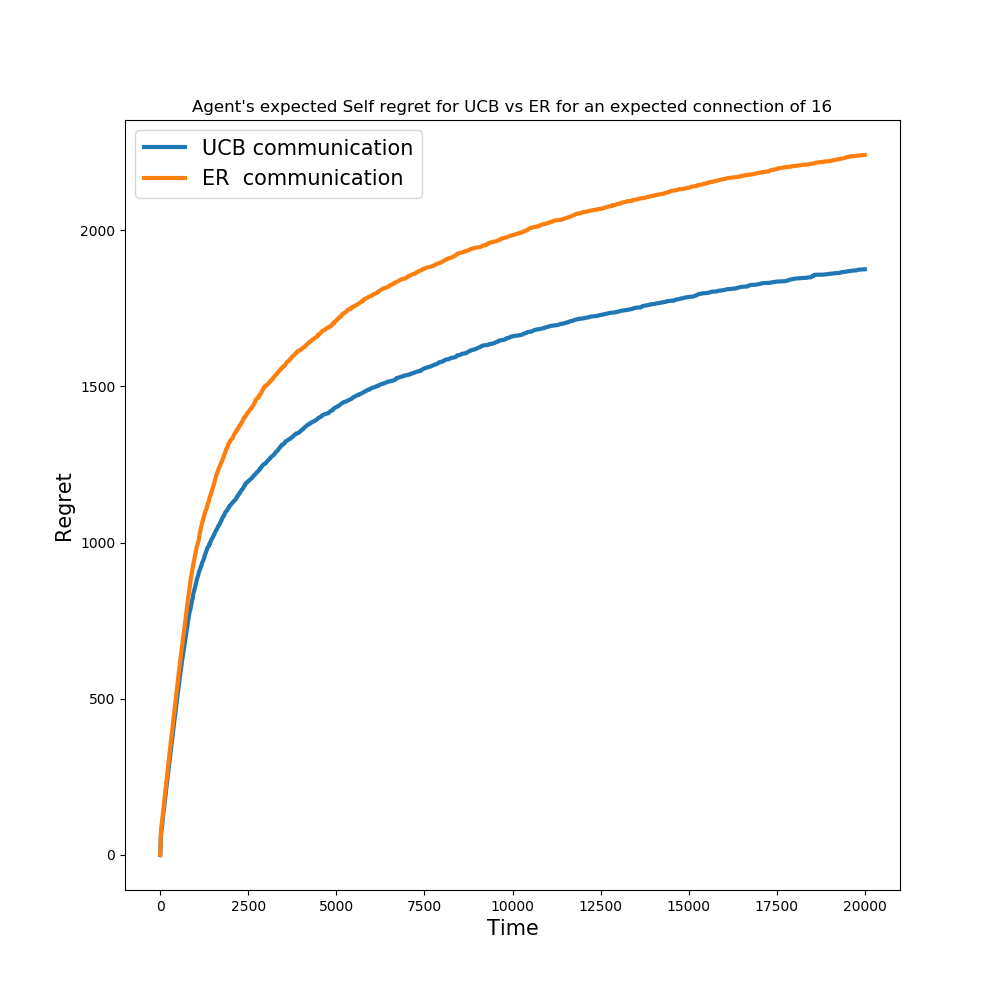}\\
	(a) Expected connectivity is 4 & (b) Expected connectivity is 16
	\end{tabular}		
	\caption{The comparison of the expected values of the self regret $R^s_{ij}(T)$ between the:  iid ER graph communication policy, UCB based communication policy (\ref{eq:UCB_Agent_allocation})
	for connectivity values of: (a) 4, and (b) 16.}	\label{fig:Comparison}
	\end{figure}

\begin{figure}[H]
	\centering	
		\includegraphics[scale = 0.315]{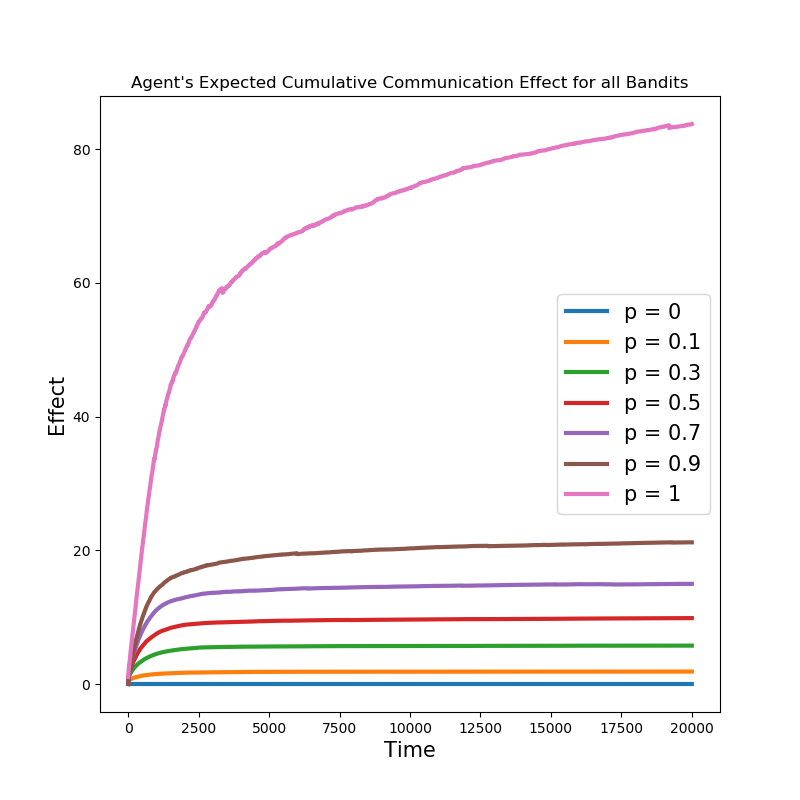}
		\includegraphics[scale = 0.315]{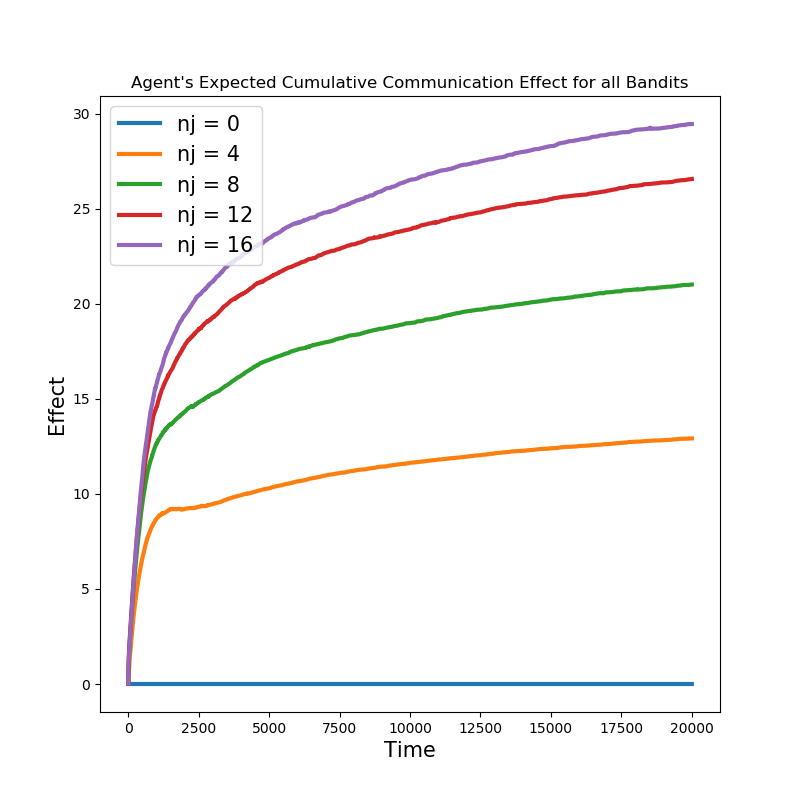}
		\caption{The communication effect $C_{j}(T)$ 
	for several connectivity values of the communication graph for the: (a) iid ER graph communication policy, (b) UCB based communication policy (\ref{eq:UCB_Agent_allocation}).}\label{fig:Comm_Effect}	
	\end{figure}

Figures \ref{fig:ERRegret} 
shows the estimates of the cumulative self regret, $R^s_{ij}(T)$, for several connectivities of the graph for the (a) the iid ER graph communication policy and  (b) the UCB based communication policy (\ref{eq:UCB_Agent_allocation}). 
Figure \ref{fig:ERRegret}demosntrates how the self regret decreases as communication is increased.
A comparison of the two simulations also clearly indicates that the dependent UCB based communication strategy significantly outperforms the independent ER graph based communication. This is further highlighted in figure \ref{fig:Comparison}. Finally figure \ref{fig:Comm_Effect} show how the communication effect increases with the expected connectivity and figure \ref{fig:Comm_EffectComparison} shows the effectiveness of the UCB based communication in comparison with the ER graph based random communication.


	\vspace{0.25cm}

\section{Conclusion}
The paper reports a novel approach for solving a class of MAMAB problems with motion and and communication constraints. The options are taken to be distributed along the nodes of a fixed graph. The agents are free to move from one node to another along the edges of the graph. They can only move one edge at a time and can only pick the option that is located at the node that it occupies. When the graph is fully connected the problem reduces to a decentralized MAMAB problem with no option availability constraints. 
This paper propose a novel decentralized policy for the motion of the agents and selection of the options that guarantee the logarithmic bounding of the regret for any communication strategy the agents might choose to employ. We investigate the effect of two communication policies for the agents: one that is independent of the information that the agents have and another that depends on the local information an agent has. The former is modeled after an iid ER graph communication strategy while the latter is based on a novel UCB type communication strategy where agents choose to communicate with others who they expect to be exploring at the given time. Using simulations we demonstrate the effectiveness of communication and the superior performance of the UCB type communication policy over the  ER graph based communication policy. The simplicity of the scheme makes the scheme scalable and very communication-ally and computationally cost effective. To the best of our knowledge it is the first time that such results have been presented.


\bibliographystyle{IEEEtran}
\bibliography{DynamicBandit}

\begin{appendix}

\subsection{Logarithmic Regret Bounds}
\addtocounter{proofoflemma}{1}
\begin{proofoflemma}
In the following we will prove the lemma \ref{Lemm:Main} by closely following the proof provided in \cite{LaiRobbins}. Let $C_{ik}^t\triangleq  \sqrt{\frac{(1+\alpha \bar{\tau})}{\left(1+\alpha\, d\left(n_k^t,r\right)\right)}\frac{\Psi_k(t)}{N_{ik}(t)}}$. Let 

For any $i\neq i_{*}$ and some $0<l<t$ define
$\mathcal{A}_{ik}^t\triangleq\{\widehat{X}_{i_*k}^t+C_{i_*k}^t\geq \widehat{\mu}_{i_*k}^t\}$, 
$\mathcal{B}_{ik}^t\triangleq\{\widehat{\mu}_{i_*k}^t\geq \widehat{\mu}_{ik}^t+2{C_{ik}^t}\}$,
$\mathcal{C}_{ik}^t\triangleq\{\widehat{\mu}_{ik}^t+2{C_{ik}^t}\geq \widehat{X}_{ik}^t+{C_{ik}^t}\}$,
where $\widehat{\mu}_{ik}^t$ is the conditional expectation of the estimate $\widehat{X}_{ik}^t$.
Then we have,
$\{\mathcal{A}_{ik}^t\cap \mathcal{B}_{ik}^t \cap \mathcal{C}_{ik}^t \}\subseteq \{Q_{i_*k}^t\geq  {Q_{ik}^t}\}$.
This implies that,
$\{Q_{i_*k}^t < {Q_{ik}^t}\} \subseteq {\bar{\mathcal{A}}_{ik}^t}\bigcup {\bar{\mathcal{B}}_{ik}^t} \bigcup {\bar{\mathcal{C}}_{ik}^t}$,
where the over bar denotes the complement of the set.

Thus for some $l>0$
${\{ Q_{i_*k}^t< {Q_{ik}^t}\:\:\&\:\: N_{ik}(t)> l\:\: \&\:\:{i \neq i_*}\}}\subseteq {\bar{\mathcal{A}}_{ik}^t}\bigcup {\widetilde{\mathcal{B}}_{ik}^t} \bigcup {\bar{\mathcal{C}}_{ik}^t}$,
where
$\widetilde{\mathcal{B}}_{ik}^t\triangleq \bar{\mathcal{B}}_{ik}^t\cap {\{ N_{ik}(t)> l\:\: \&\:\:{i \neq i_*}\}}$.

From the above expressions we have,
{\small
\begin{align*}
\mathcal{P}({\{Q_{i_*k}^t< {Q_{ik}^t}\:\:\&\:\: N_{ik}(t)> l\:\: \&\:\:{i \neq i_*}\}})&\leq \mathcal{P}({\bar{\mathcal{A}}_{ik}^t})+\mathcal{P}({\widetilde{\mathcal{B}}_{ik}^t})+\mathcal{P}({\bar{\mathcal{C}}_{ik}^t}).
\end{align*}
}
and hence that
\begin{align*}
\mathcal{P}({\{Q_{i_*k}^t< {Q_{ik}^t}\:\:\&\:\: N_{ik}(t)> l\:\: \&\:\:{i \neq i_*}\}})&\leq \mathcal{P}(\widetilde{\mathcal{B}}^t_{ik})+2\max_{r}\mathcal{P}(\{|\widehat{X}_{rk}^t -\widehat{\mu}_{rk}^t|>C_{sk}^t\}).
\end{align*}

What remains to complete the bound is to find an upper bound for $\widetilde{\mathcal{B}}^t_{ik}$. \emph{We consider the case where there exists a well defined optimal arm at all times. That is the case where there exists a $\Delta, \bar{\Delta}>0$ such that $\Delta\leq E(X_{i_*}^\nu )-E(X_i^\nu )\leq \bar{\Delta}$ for all $r,s>0$.}
Then 
since $\bar{\mathcal{B}}^t_{ik}=\{\widehat{\mu}_{i_*k}^t<\widehat{\mu}_{ik}^t+2{C^t_{ik}}\}$
\begin{align*}
\bar{\mathcal{B}}^t_{ik}&\subseteq
\left\{\frac{\Delta}{2}< \sqrt{\frac{(1+\alpha \bar{\tau})}{\left(1+\alpha\, d\left(n_k^t,i\right)\right)}\frac{\Psi(t)}{N_{ik}(t)}}\right\}\subseteq
\left\{\frac{\Delta}{2}< \sqrt{(1+\alpha \bar{\tau})\frac{\Psi(t)}{N_{ik}(t)}}\right\},
\end{align*}
where $\Psi(t)$ is such that $\Psi_k(t)\leq \Psi(t)$ for all $k$.

Thus since $\widetilde{\mathcal{B}}^t_{ik}\subseteq \bar{\mathcal{B}}_{ik}^t\cap \left\{ N_{ik}^s(t)> l\:\:\&\:\: i\neq i_*\right\}$
we have that
{
\begin{align*}
&\mathcal{P}(\widetilde{\mathcal{B}}^t_{ik})
\leq\mathcal{P}\left(\left\{N_{ik}(t)<\frac{4(1+\alpha \bar{\tau})}{{\Delta^2}}\,\Psi(t)\:\:\&\:\: N_{ik}(t)> l\right\}\right).
\end{align*}
}

Since when 
\begin{align*}
l(t)&\triangleq\ceil*{\frac{4(1+\alpha \bar{\tau})}{{\Delta^2}}\,\Psi(t)},
\end{align*}
\begin{align*}
\mathcal{P}(\widetilde{\mathcal{B}}^t_{ik})
\leq\mathcal{P}\left(\left\{N_{ik}(t)< \frac{4(1+\alpha \bar{\tau})}{{\Delta^2}}\,\Psi(t)\:\:\&\:\: N_{ik}(t)>\, \ceil*{\frac{4(1+\alpha \bar{\tau})}{{\Delta^2}}\,\Psi(t)}\right\}\right)=0,
\end{align*}
it follows that
for any $i\neq i_*$
\begin{align*}
\mathcal{P}\left({\left\{Q_{i_*k}^t< {Q_{ik}^t}\:\:\&\:\: N_{ik}(t)> \ceil*{\frac{4(1+\alpha \bar{\tau})}{{\Delta^2}}\,\Psi(t)}\:\: \&\:\:{i \neq i_*}\right\}}\right)\leq 2\max_{r}\mathcal{P}\left(\left\{\left|\widehat{X}_{rk}^t- \widehat{\mu}_{rk}^t\right|>\sqrt{\frac{(1+\alpha \bar{\tau})}{\left(1+\alpha\, d\left(n_k^t,r\right)\right)}\frac{\Psi_{k}(t)}{N_{rk}(t)}}\right\}\right).
\end{align*}


From (\ref{eq:GarivierHoeffding}) it follows that the sample mean estimator defined above satisfies the following tail bound
\begin{align*}
\mathcal{P}\left(\left\{\left|\widehat{X}_{rk}^t-\widehat{\mu}_{rk}^t\right|>\sqrt{\frac{\Psi(t)}{N_{rk}(t)}}\right\}\right)\leq  
\frac{2\vartheta}{t^2},
\end{align*}
where $\Psi(t)$ satisfies $({\sigma}^2\sqrt{1+\eta}))\log{\left(t\sqrt{\log{(t)}}\right)}\leq  \Psi{(t)}$ with $\vartheta=1/\log (1+\eta)$ for all $t>0$ and some $\eta>0$. 
The from the fact 
\begin{align*}
\left\{\left|\widehat{X}_{rk}^t- \widehat{\mu}_{rk}^t\right|>\sqrt{\frac{(1+\alpha \bar{\tau})}{\left(1+\alpha\, d\left(n_k^t,r\right)\right)}\frac{\Psi_{k}(t)}{N_{rk}(t)}}\right\}\subseteq
\left\{\left|\widehat{X}_{rk}^t- \widehat{\mu}_{rk}^t\right|>\sqrt{\frac{\Psi_{k}(t)}{N_{rk}(t)}}\right\}.
\end{align*}
it follows that
{
\begin{align*}
\mathcal{P}\left(\left\{Q_{i_*k}^{t}<{Q_{ik}^{t}}\right\}\right)
&=\left(\mathcal{P}\left(\left\{Q_{i_*k}^{t}<{Q_{ik}^{t}}\:\:\&\:\:N_{ik}(t)\leq\,l(t)\right\}\right)+\mathcal{P}\left(\left\{Q_{i_*k}^{t}<{Q_{ik}^{t}}\:\:\&\:\:N_{ik}(t)>\,l(t)\right\}\right)\right)\\
&\leq \frac{4\vartheta}{t^2}+\mathcal{P}\left(\left\{N_{ik}(t)\leq\,l(t)\right\}\right),
\end{align*}
}
where 
\begin{align*}
l(t)&\triangleq\ceil*{\frac{4(1+\alpha \bar{\tau})}{{\Delta^2}}\,\Psi(t)}.
\end{align*}
\end{proofoflemma}



\begin{proofoftheorem}
We begin the regret analysis by recalling that self regret, of agent $j$ due to sampling of the non optimal arm $i$ satisfies
$R_{ik}^s(T)\leq \bar{\Delta}\sum_{t=1}^T\mathcal{P}\left(\{\varphi_{k}^t=i\}\right)$.

This along with lemma \ref{Lemm:Main} gives us that 
{\small
\begin{align*}
\sum_{t=1}^T\mathcal{P}\left(\{\varphi_{k}^t=i\}\right)&\leq 2+\sum_{t=3}^T\frac{4\vartheta}{\left(t-1\right)^2}+\sum_{t=3}^T\mathcal{P}\left(\left\{Q_{i_*k}^{t-1}<{Q_{ik}^{t-1}}\:\:\&\:\: d(n_k^{t-1},i)=1\:\&\:N_{ik}(t-1)\leq\,l(t-1)\right\}\right),\\
&\leq 2+4\vartheta+\mathbb{E}\left(\sum_{t=2}^T\mathbb{I}_{\left\{N_{ik}(t-1)\leq\,l(t-1)\right\}}\right)\\
&< 2+4\vartheta+\left(l(T-1)-2\right)\frac{\mathbb{E}\left(\sum_{t=2}^T\mathbb{I}_{\left\{N_{ik}(t-1)\leq\,l(t-1)\right\}}\right)}{\mathbb{E}\left(\sum_{t=2}^T\mathbb{I}_{\left\{N_{ik}^s(t-1)\leq\,l(t-1)\right\}}\right)}\leq 4\vartheta+l(T).
\end{align*}
}
Thus we have
$
R^s_{ik}(T)<\bar{\Delta}\left(2(1-f_{ik}(T))+4\vartheta+f_{ik}(T)\,l(T)\right)\leq 4\vartheta+l(T)$,
where
\begin{align*}
f_{ik}(T)\triangleq \frac{\mathbb{E}\left(\sum_{t=2}^T\mathbb{I}_{\left\{N_{ik}(t-1)\leq\,l(t-1)\right\}}\right)}{\mathbb{E}\left(\sum_{t=2}^T\mathbb{I}_{\left\{N^s_{ik}(t-1)\leq\,l(t-1)\right\}}\right)}.
\end{align*}

When there is no communication it is clear that $N_{ik}(t)= N^s_{ik}(t)$ and thus that $f_{ik}(T)=1$ when there is no communication. 
Since $N_{ik}(t)\geq N^s_{ik}(t)$, with equality holding when there is no communication, it follows that $\left\{N_{ik}(t)\leq\,l(t)\right\}\subseteq \left\{N^s_{ik}(t)\leq\,l(t)\right\}$. This implies that $f_{ik}(T)\leq 1$ with equality holding when there is no communication. It also shows that $f_{ik}(T)$ reduces with increasing communication.

\end{proofoftheorem}

\begin{proofoflemma}
Let $\mathcal{N}_{jP}$ be the space of all subsets of $\left\{1,2,\cdots,n_A\right\}$ that contain $j$. The discrete random variable $\mathcal{N}_j^t$ takes values in the set $\mathcal{N}_{jP}$. Then we see that for any $i\neq i_*$
\begin{align*}
\sum_{t=1}^T\mathcal{P}\left(\{\epsilon_{ij}^t=1\}\}\right)
&=  \sum_{t=1}^T\mathcal{P}\left(\cup_{k\in \mathcal{N}_{j}^t}\left\{\varphi_{k}^t=i\right\}\right)
\leq \sum_{t=1}^T\sum_{\mathcal{N}_{j\alpha}\in \mathcal{N}_{jP}}\sum_{k\in \mathcal{N}}\mathcal{P}\left(\{\varphi_{k}^t=i\} \cap\{\mathcal{N}_j^t=\mathcal{N}_{j\alpha}\}\right)\\
&= \sum_{t=1}^T\sum_{\mathcal{N}_{j\alpha}\in \mathcal{N}_{jP}}\sum_{k\in \mathcal{N}}\mathcal{P}\left(\{\mathcal{N}_j^t=\mathcal{N}_{j\alpha}\}\,|\,\{\varphi_{k}^t=i\} \right)\mathcal{P}\left(\{\varphi_{k}^t=i\} \right)
\end{align*}
Thus we have
\begin{align*}
\mathbb{E}(N_{ij}(T))
&\leq \sum_{\mathcal{N}_{j\alpha}\in \mathcal{N}_{jP}}\max_{k,t\leq T}\mathcal{P}\left(\{\mathcal{N}_j^t=\mathcal{N}_{j\alpha}\}\,|\,\{\varphi_{k}^t=i\} \right)\sum_{t=1}^T\sum_{\stackrel{k\in \mathcal{N}_{j\alpha}}{k\neq j}}\mathcal{P}\left(\{\varphi_{k}^t=i\} \right)\\
&\leq \left(\max_k\mathbb{E}\left(N^s_{ik}(T)\right)\right)\left(\max_{k,t\leq T}\left\langle |\mathcal{N}_{j}^t|\right\rangle_{\mathcal{P}\left(\mathcal{N}_j^t\,|\,\varphi_{k}^t=i \right)}\right)
\leq \left(4\vartheta+\,l(T)\right)\left(\max_{k,t\leq T}\left\langle |\mathcal{N}_{j}^t|\right\rangle_{\mathcal{P}\left(\mathcal{N}_j^t\,|\,\varphi_{k}^t=i \right)}\right)
\end{align*}
where we have defined
\begin{align*}
\left\langle |\mathcal{N}_{j}^t|\right\rangle_{\mathcal{P}\left(\mathcal{N}_j^t\,|\,\varphi_{k}^t=i \right)} 
&\triangleq\sum_{\mathcal{N}_{j\alpha}\in \mathcal{N}_{jP}}\mathcal{P}\left(\{\mathcal{N}_j^t=\mathcal{N}_{j\alpha}\}\,|\,\{\varphi_{k}^t=i\} \right)\left(|\mathcal{N}_{j\alpha}|\right),
\end{align*}
\end{proofoflemma}


\end{appendix}
\end{document}